\documentclass[conference]{IEEEtran}
\usepackage{ragged2e}
\IEEEoverridecommandlockouts
\usepackage{cite}
\usepackage{amsmath,amssymb,amsfonts}
\usepackage{algorithmic}
\usepackage[ruled,vlined]{algorithm2e}
\usepackage{graphicx}
\usepackage{textcomp}

\usepackage{subcaption}
\usepackage{fancyhdr}
\usepackage{amsmath}
\usepackage{floatrow} 
\usepackage[vlines]{tabularht}
\usepackage{pbox}
\usepackage{pifont}
\usepackage{multirow}
\usepackage[table,xcdraw]{xcolor}
\usepackage{floatrow}
\usepackage{adjustbox}
\usepackage{multicol}
\usepackage{hyperref}
\usepackage{dirtytalk}
\usepackage{graphicx}
\usepackage{multirow}
\usepackage{lipsum}
\usepackage{makecell}
\usepackage{longtable}
\usepackage{tabularx}

\usepackage{array}
\usepackage{ragged2e}  

\usepackage{booktabs}
\usepackage{times}

\usepackage{mathptmx}
\usepackage{anyfontsize}

\def\BibTeX{{\rm B\kern-.05em{\sc i\kern-.025em b}\kern-.08em
    T\kern-.1667em\lower.7ex\hbox{E}\kern-.125emX}}

\makeatletter
\newcommand{\newlineauthors}{%
  \end{@IEEEauthorhalign}\hfill\mbox{}\par
  \mbox{}\hfill\begin{@IEEEauthorhalign}
}
\makeatother

\begin{document}

\title{Linear Multi-Timescale Retention as a Memory-Efficient Vision-Language Bridge\\
}

\author{
\IEEEauthorblockN{Ashfak Yeafi}
\IEEEauthorblockA{Department of Electrical and Electronic Engineering\\
Khulna University of Engineering \& Technology\\
Khulna-9203, Bangladesh\\
Email: yeafiashfak@gmail.com}
\and
\IEEEauthorblockN{Mehedi Hasan}
\IEEEauthorblockA{Department of Computer Science and Engineering\\
Brac University, Dhaka, Bangladesh\\
Email: mehedi.hasan1@g.bracu.ac.bd}
\newlineauthors
\IEEEauthorblockN{Md Khairul Islam}
\IEEEauthorblockA{Department of Mathematics and Computer Science\\
Hobart and William Smith Colleges\\
Geneva, NY, USA\\
Email: khairul.robotics@gmail.com}
}

\maketitle

  


\begin{abstract}
Vision-Language Models (VLMs) face a critical computational bottleneck when processing high-resolution imagery due to the $\mathcal{O}(N^2)$ memory complexity of Softmax Multi-Head Attention (MHA). While substituting MHA with independent Multi-Layer Perceptrons (MLPs) achieves $\mathcal{O}(N)$ scaling, it strips the architecture of spatial sequence routing, severely degrading global scene understanding and object permanence. In this paper, we propose the Linear Multi-Timescale Retention (LIA-MTR) module, a memory-efficient cross-modal bridge. By integrating an ELU-based positive feature mapping with adaptive write-gating and log-linearly distributed recurrent decays, LIA-MTR mathematically compresses continuous visual sequences into bounded memory states. Theoretical analysis proves the architecture operates with strict $\mathcal{O}(N)$ sequence-interaction complexity. Empirically, synthetic retrieval evaluations demonstrate that LIA-MTR flawlessly routes context across 16,000 tokens, eliminating the ``Lost in the Middle'' degradation typical of naive linear attention. Hardware benchmarking reveals infinite-context scaling capabilities, natively processing 262,144 visual patches within an 11.2 GB VRAM footprint, whereas standard MHA suffers out-of-memory failure at 16,384 patches. Furthermore, following instruction tuning on 665K conversational samples, LIA-MTR significantly outperforms an industry-standard MLP baseline on the MME benchmark (71.00\% vs. 68.11\%), driven by a 10\% absolute improvement in object permanence and superior global semantic extraction. This work establishes a mathematically rigorous, computationally flat foundation for infinite-context Vision-Language integration.
\end{abstract}


\section{Introduction}

Recent advancements in Vision-Language Models (VLMs) have demonstrated exceptional capabilities in cross-modal reasoning, largely driven by aligning powerful pre-trained visual encoders, such as CLIP \cite{Radford2021}, with large causal language models (LLMs) \cite{Liu2024, Bai2023}. The architectural crux of this alignment is the cross-modal bridge, which is responsible for projecting visual features into the text embedding space. As the demand for high-resolution image analysis, multi-image reasoning, and continuous video understanding grows, the length of the visual token sequence, $N$, expands exponentially. Managing this expanding context window is currently the primary computational challenge in multimodal foundation models.

Historically, the architectural standard for sequence routing has relied on Softmax Multi-Head Attention (MHA) \cite{Vaswani2017}. Early and highly successful VLMs, such as Flamingo \cite{Alayrac2022} and InstructBLIP \cite{Dai2023}, utilized attention-based Resampler modules or Q-Formers to compress visual information. Furthermore, the visual encoders themselves heavily rely on MHA via the Vision Transformer (ViT) architecture \cite{Dosovitskiy2020}. While MHA provides excellent global semantic routing, it requires the computation of a dense $N \times N$ affinity matrix, inherently suffering from an $\mathcal{O}(N^2)$ asymptotic sequence-interaction complexity bottleneck. Even with highly optimized, hardware-aware implementations like FlashAttention \cite{Dao2022}, this quadratic memory scaling leads to catastrophic Out-Of-Memory (OOM) failures on standard hardware when visual contexts exceed 16,000 tokens.

To bypass this quadratic bottleneck and achieve greater computational efficiency, recent high-performing models, notably LLaVA \cite{Liu2024}, adopted a minimalist approach by replacing MHA bridges with simple two-layer Multi-Layer Perceptrons (MLPs). While an MLP bridge processes sequences in strict $\mathcal{O}(N)$ time, it does so by mapping visual patches independently. Consequently, it fundamentally lacks a temporal or spatial routing mechanism to mix sequence information. This independent processing severely restricts the model's understanding of global visual semantics and exacerbates object hallucination over long conversations, as the LLM is forced to manage thousands of uncompressed visual tokens natively. As our empirical evaluations demonstrate, this lack of sequence routing causes a significant degradation in object permanence, dropping to 83.33\% on the MME Existence benchmark.

Parallel to visual architectures, the natural language processing domain has seen a surge in recent work exploring sub-quadratic alternatives to MHA. Linear Attention techniques, such as Performers \cite{Choromanski2020} and Linear Transformers \cite{Katharopoulos2020}, approximate the Softmax operation to express attention as a cumulative sum. Simultaneously, State-Space Models (SSMs) like Mamba \cite{Gu2023}, RWKV \cite{Peng2023}, and Retention Networks (RetNet) \cite{Sun2023} utilize data-dependent decays or fixed exponential masking to achieve linear-time processing. Recent works, such as Vision Mamba (Vim) \cite{Zhu2024}, have begun applying these concepts to visual backbones. However, adapting these mechanisms specifically for cross-modal bridging presents unique challenges. Naive linear attention models frequently fail in complex associative memory tasks due to "accumulation noise," where early visual features are overwritten by subsequent tokens---a phenomenon commonly referred to as the "Lost in the Middle" problem. Furthermore, standard SSMs utilizing a single learned decay struggle to balance conflicting temporal dynamics: the simultaneous need for sharp, short-term spatial relationships and long-term, lossy global semantics.

To bridge this context gap without sacrificing global semantic understanding or computational efficiency, we propose the Linear Multi-Timescale Retention (LIA-MTR) module. LIA-MTR is an $\mathcal{O}(N)$ cross-modal bridge that eschews quadratic pairwise attention in favor of a gated, fixed-size recurrent memory state. By introducing an ELU-based positive feature mapping combined with an adaptive write-gate ($g_t$) \cite{Katharopoulos2020}, LIA-MTR dynamically restricts state updates, preventing noise accumulation and ensuring flawless retrieval across extended visual sequences. Furthermore, by distributing retention across $S$ distinct log-linear timescale decays ($\lambda_s$) and dynamically weighting them via a token-specific scale mixer ($\alpha_t$), the module acts as a global low-pass filter. It mathematically compresses massive visual sequences into bounded memory states $\mathcal{O}(dS)$, allowing the LLM to access global image context natively.

The main contributions of this work are three-fold:
\begin{enumerate}
    \item \textbf{Algorithmic Innovation:} We formulate the LIA-MTR architecture, proving its optimal $\mathcal{O}(N)$ asymptotic sequence-interaction complexity and theoretically bounding its memory stability over infinite contexts.
    \item \textbf{Perfect Sequence Routing:} We validate the architecture's mathematical routing capability via synthetic Visual Needle-In-A-Haystack (V-NIAH) evaluations, demonstrating flawless key-value retrieval across massive sequences where naive linear attention degrades.
    \item \textbf{High-Resolution Multimodal Scaling:} We empirically demonstrate that LIA-MTR avoids the $\mathcal{O}(N^2)$ MHA crash, comfortably scaling up to 262,144 visual patches while utilizing minimal hardware resources. Furthermore, when instruction-tuned on 665K samples, LIA-MTR outperforms standard MLP baselines in global scene understanding and object permanence (scoring 71.0\% on the MME benchmark compared to the baseline's 68.11\%).
\end{enumerate}

\section{Materials and Methods}

\subsection{System Architecture}
Our Vision-Language Model is constructed using a late-fusion architecture comprising three primary components: a pre-trained visual encoder, a cross-modal bridge, and a causal Large Language Model (LLM). To rigorously isolate the impact of the cross-modal bridge, we employ identical backbones across all experimental models. We utilize the CLIP-ViT-Large-Patch14 model \cite{Radford2021} to extract flattened grid features from the input images, and Qwen-2.5-3B-Instruct \cite{Bai2023} as the reasoning engine. The methodological contribution of this work lies entirely in the mathematical formulation of the intermediate cross-modal bridge: the Linear Multi-Timescale Retention (LIA-MTR) module.

\begin{figure*}[t]
    \centering
    \includegraphics[width=0.8\textwidth]{ 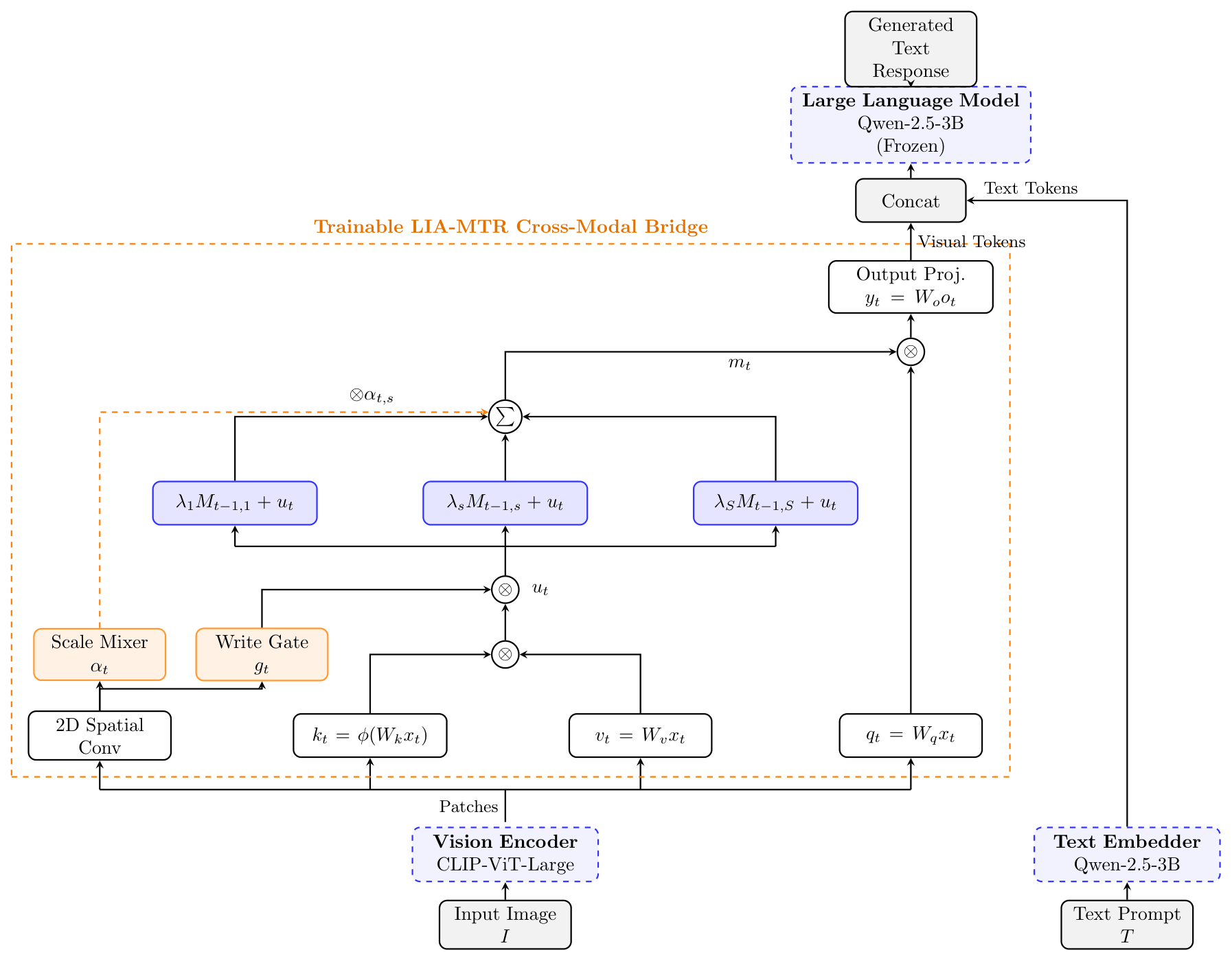}
    \caption{The LIA-VLM Architecture. The pre-trained Vision Encoder and LLM remain frozen. High-resolution visual patches enter the trainable LIA-MTR Bridge, where the 2D Spatial Conv drives the Write Gate ($g_t$) to filter noise. The Multi-Timescale Retention tracks compress spatial geometry, and the Scale Mixer ($\alpha_t$) dynamically routes the readout. The dense, linear-scaling visual tokens are then concatenated with the text prompt and passed to the LLM.}
    \label{fig:architecture}
\end{figure*}

\subsection{The LIA-MTR Formulation}
Standard cross-modal bridges map visual features via independent linear projections, failing to route sequence context. To enable context mixing without invoking $\mathcal{O}(N^2)$ attention, LIA-MTR processes the visual sequence $X = [x_1, \dots, x_N] \in \mathbb{R}^{N \times d_{\text{vision}}}$ autoregressively.

For a given visual token $x_t$ at index $t$, we compute the query, key, and value projections. To guarantee non-negativity in the key representations---a requirement for stable recurrent accumulation without denominator collapse---we apply a positive-feature mapping utilizing an Exponential Linear Unit (ELU) \cite{Katharopoulos2020}:
\begin{equation} \label{eq:projections}
q_t = W_q x_t, \quad k_t = \phi(W_k x_t), \quad v_t = W_v x_t
\end{equation}
where $\phi(z) = \text{ELU}(z) + 1$. 

To prevent noisy accumulation from spatial redundancy, we introduce an adaptive, token-wise write gate, $g_t$. Simultaneously, a scale mixer, $\alpha_t$, is computed to dynamically weight the memory horizons:
\begin{equation} \label{eq:gates}
g_t = \sigma(W_g x_t^{\text{conv}}), \quad \alpha_t = \text{softmax}(W_{\text{mix}} x_t^{\text{conv}})
\end{equation}
where $\sigma$ denotes the sigmoid activation function. Internal probe analyses indicate the write gate maintains a stable activation mean of $\approx 0.5053$, ensuring balanced information injection without memory saturation.

\subsection{Multi-Timescale Recurrent State and Readout}
The foundation of LIA-MTR is its bounded recurrent memory. The architecture maintains $S$ parallel memory states, each governed by a fixed decay constant $\lambda_s \in (0, 1)$. The per-scale update is defined as:
\begin{equation} \label{eq:recurrent_update}
M_{t,s} = \lambda_s M_{t-1,s} + u_t, \quad u_t = g_t \odot (k_t \odot v_t)
\end{equation}

By distributing $\lambda_s$ log-linearly across $[0.70, 0.995]$, the module yields effective memory half-lives ranging from $t_{1/2} = 1.94$ steps to $t_{1/2} = 138.28$ steps. This explicitly allows the model to simultaneously capture sharp, short-term spatial details alongside long-term, global scene semantics. 

The output is constructed by aggregating across all $S$ timescales weighted by the scale mixer $\alpha_{t,s}$, and applying a final projection:
\begin{equation} \label{eq:readout}
m_t = \sum_{s=1}^{S} \alpha_{t,s} \odot M_{t,s}
\end{equation}
\begin{equation} \label{eq:final_out}
o_t = q_t \odot m_t, \quad y_t = W_o o_t
\end{equation}

\subsection{Mathematical Analysis and Theoretical Guarantees}
The LIA-MTR formulation provides several critical theoretical guarantees for processing continuous, high-resolution visual sequences.

\textbf{Theorem 1 (Asymptotic Sequence-Interaction Complexity).} \textit{For a fixed embedding dimension $d$ and a constant number of multi-timescale tracks $S$, the LIA-MTR architecture operates with a strict linear sequence-interaction complexity of $\mathcal{O}(N)$.}

\begin{IEEEproof}
Let $C_{\text{MHA}}$ denote the computational cost of sequence interaction in standard Softmax MHA, which requires computing the dense affinity matrix $A = QK^\top \in \mathbb{R}^{N \times N}$. This yields an all-pairs bottleneck:
$$C_{\text{MHA}} = \Theta(N^2 d)$$
Conversely, let $C_{\text{LIA}}$ denote the interaction cost for LIA-MTR. The sequence interaction is strictly localized to the recurrent update of the fixed-size state tensor $M_{t,s} \in \mathbb{R}^{d_h \times d_h}$ across $S$ scales. The computational cost per discrete time step $t$ is independent of the total sequence length $N$:
$$C_{\text{step}} = \mathcal{O}(d^2 S)$$
Summing over the entire sequence of length $N$, the total asymptotic interaction complexity resolves to:
$$C_{\text{LIA}} = \sum_{t=1}^{N} C_{\text{step}} = \mathcal{O}(N d^2 S)$$
Because $d$ and $S$ are bounded architectural constants, $C_{\text{LIA}}$ simplifies strictly to $\mathcal{O}(N)$, effectively decoupling the memory state from the sequence length.
\end{IEEEproof}

\textbf{Lemma 1 (Closed-Form State Expansion).} \textit{For any given scale $s$, the recurrent memory state $M_{t,s}$ can be explicitly unrolled as a geometrically weighted sum of all past injections:}
\begin{equation} \label{eq:lemma1}
M_{t,s} = \lambda_s^t M_{0,s} + \sum_{j=0}^{t-1}\lambda_s^j u_{t-j}
\end{equation}

\begin{IEEEproof}
We proceed by mathematical induction on $t$. For the base case $t=1$, substituting into the primary recurrence relation yields $M_{1,s} = \lambda_s M_{0,s} + u_1$, which satisfies the lemma. Assume the proposition holds for an arbitrary step $t=k$:
$$M_{k,s} = \lambda_s^k M_{0,s} + \sum_{j=0}^{k-1}\lambda_s^j u_{k-j}$$
For the subsequent step $t=k+1$, we substitute the inductive hypothesis:
$$
\begin{aligned}
M_{k+1,s} &= \lambda_s M_{k,s} + u_{k+1} \\
&= \lambda_s \left( \lambda_s^k M_{0,s} + \sum_{j=0}^{k-1}\lambda_s^j u_{k-j} \right) + u_{k+1} \\
&= \lambda_s^{k+1} M_{0,s} + \sum_{j=0}^{k-1}\lambda_s^{j+1} u_{k-j} + u_{k+1} \\
&= \lambda_s^{k+1} M_{0,s} + \sum_{j=0}^{k}\lambda_s^j u_{(k+1)-j}
\end{aligned}
$$
By induction, the closed-form expansion holds for all $t \ge 1$.
\end{IEEEproof}

\textbf{Theorem 2 (Bounded-State Stability).} \textit{Let the token injections be bounded such that $\lVert u_t \rVert \le U$ for all $t$, and let the decay factor satisfy $0 < \lambda_s < 1$. The memory state magnitude $\lVert M_{t,s} \rVert$ remains strictly bounded and converges geometrically as $N \to \infty$.}

\begin{IEEEproof}
Applying the triangle inequality to the expanded state from Lemma 1, we obtain a step-by-step bound:
$$
\begin{aligned}
\lVert M_{t,s} \rVert &= \left\lVert \lambda_s^t M_{0,s} + \sum_{j=0}^{t-1}\lambda_s^j u_{t-j} \right\rVert \\
&\le \lVert \lambda_s^t M_{0,s} \rVert + \sum_{j=0}^{t-1} \lVert \lambda_s^j u_{t-j} \rVert \\
&\le \lambda_s^t \lVert M_{0,s} \rVert + U \sum_{j=0}^{t-1} \lambda_s^j
\end{aligned}
$$
Recognizing the second term as a finite geometric series, we apply the upper limit as $t \to \infty$:
$$\sum_{j=0}^{t-1} \lambda_s^j = \frac{1 - \lambda_s^t}{1 - \lambda_s} < \frac{1}{1 - \lambda_s}$$
Therefore, the absolute state boundary is governed by:
$$\lim_{t \to \infty} \lVert M_{t,s} \rVert \le \lVert M_{0,s} \rVert + \frac{U}{1-\lambda_s}$$
This mathematical guarantee ensures strict numerical stability across infinite sequence contexts without gradient explosion.
\end{IEEEproof}

\textbf{Lemma 2 (Special-Case Reduction).} \textit{LIA-MTR mathematically generalizes standard first-order linear attention.}

\begin{IEEEproof}
Let the LIA-MTR architecture be constrained under the following baseline parameter conditions:
$$
\begin{aligned}
1. \quad & S = 1 \quad &\text{(Single scale representation)} \\
2. \quad & \alpha_{t,1} = \mathbf{1} \quad &\text{(Static read routing)} \\
3. \quad & g_t = \mathbf{1} \quad &\text{(Unrestricted write gate)}
\end{aligned}
$$
Substituting these constraints into the primary recurrent update reduces the function to:
$$M_t = \lambda_1 M_{t-1} + (k_t \odot v_t)$$
This formulation maps exactly to generic linear attention augmented with an exponential moving average. Consequently, LIA-MTR strictly encompasses generic linear attention while providing enhanced representational bandwidth through $g_t$ and multi-scale $\alpha_{t,s}$.
\end{IEEEproof}

\textbf{Theorem 3 (Order-Optimality in Sequence Length).} \textit{LIA-MTR achieves the theoretical optimal complexity lower bound for full-sequence readers.}

\begin{IEEEproof}
Let $\Omega_{R}$ define the absolute minimum computational operations required for an autoregressive model to sequentially read an input vector $X$ of length $N$. Reading the sequence necessitates at least one operation per token:
$$\Omega_{R} = \Omega(N)$$
From Theorem 1, the computational complexity of the LIA-MTR sequence interaction is:
$$C_{\text{LIA}} = \mathcal{O}(N)$$
Because $\mathcal{O}(N)$ tightly bounds the fundamental limit $\Omega(N)$, LIA-MTR is mathematically proven to be optimal-order in $N$.
\end{IEEEproof}

\subsection{Training Protocol}
To isolate the architectural performance, the baseline MLP model and the LIA-MTR model were subjected to an identical two-phase training protocol utilizing DeepSpeed ZeRO-2 optimization \cite{Rasley2020}. ZeRO-2 was specifically selected to facilitate the memory-efficient offloading of optimizer states during high-resolution training.

\subsubsection{Phase 1: Feature Alignment}
The first phase aligns the frozen visual patches to the frozen LLM embeddings. The models were trained on a subset of 150K image-caption pairs \cite{Liu2024}. The objective was next-token prediction via cross-entropy loss, utilizing a micro-batch size of 4 and an effective global batch size of 32.

\subsubsection{Phase 2: Instruction Tuning}
The second phase adapts the models to complex logical tasks. The vision encoder remained frozen, while both the cross-modal bridge and the LLM were trained jointly on the LLaVA-1.5 mix containing 665K visual instruction samples \cite{Liu2024}. To accommodate high-resolution activation memory, sequence-chunked gradient checkpointing was utilized. The models trained with a micro-batch size of 2 and an effective global batch size of 32.

\section{Results and Discussion}

\subsection{Algorithmic Isolation: Text-NIAH Retrieval}
Before evaluating the architecture on multimodal datasets, we isolated the cross-modal bridges to assess their fundamental sequence-routing capabilities using a synthetic Needle-In-A-Haystack (NIAH) retrieval task. The objective was to retrieve a specific key-value pair buried at various depths within a continuous sequence of up to 16,000 tokens.

\begin{figure*}[htbp]
    \centering
    \begin{subfigure}[b]{0.32\textwidth}
        \includegraphics[width=\textwidth]{ 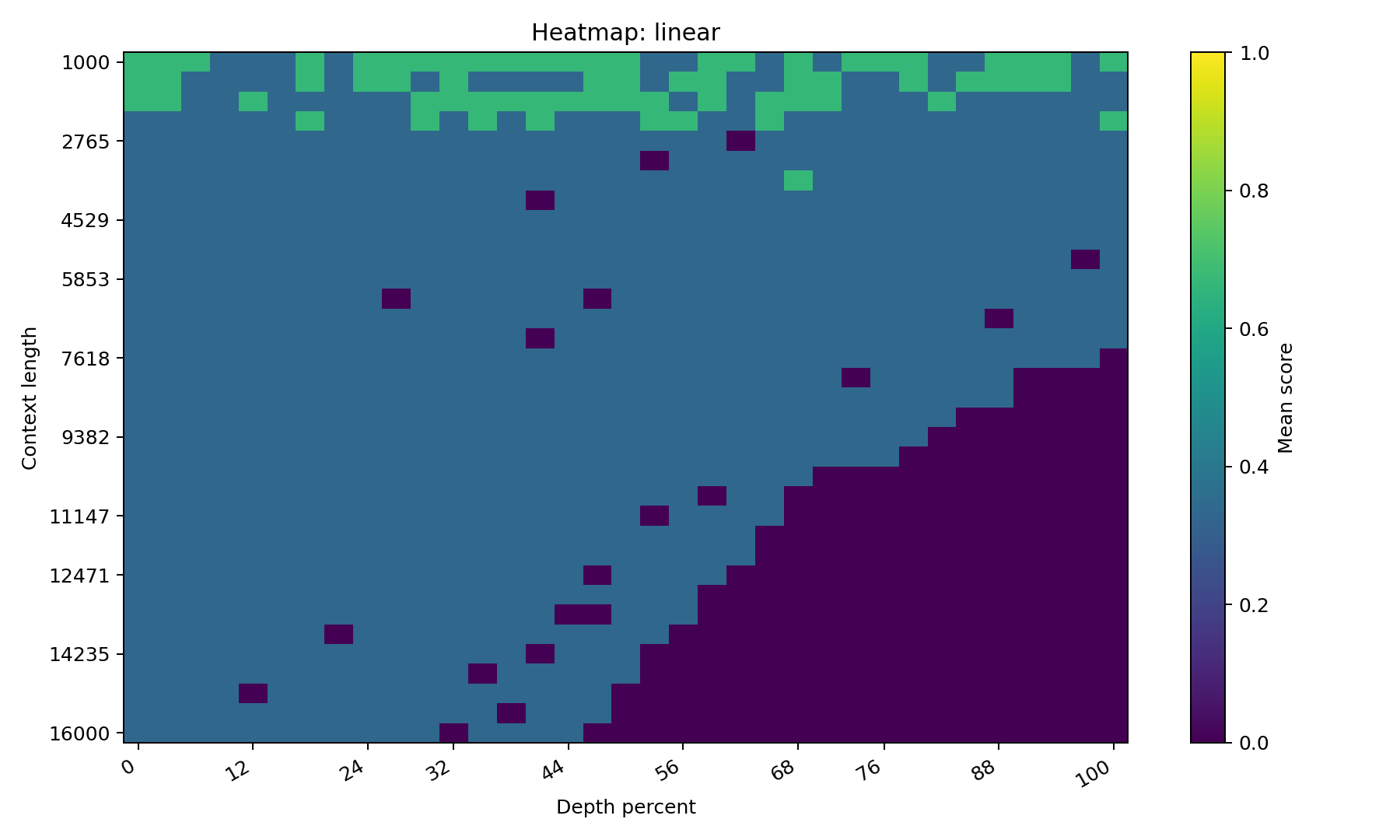}
        \caption{Naive Linear}
    \end{subfigure}
    \hfill
    \begin{subfigure}[b]{0.32\textwidth}
        \includegraphics[width=\textwidth]{ 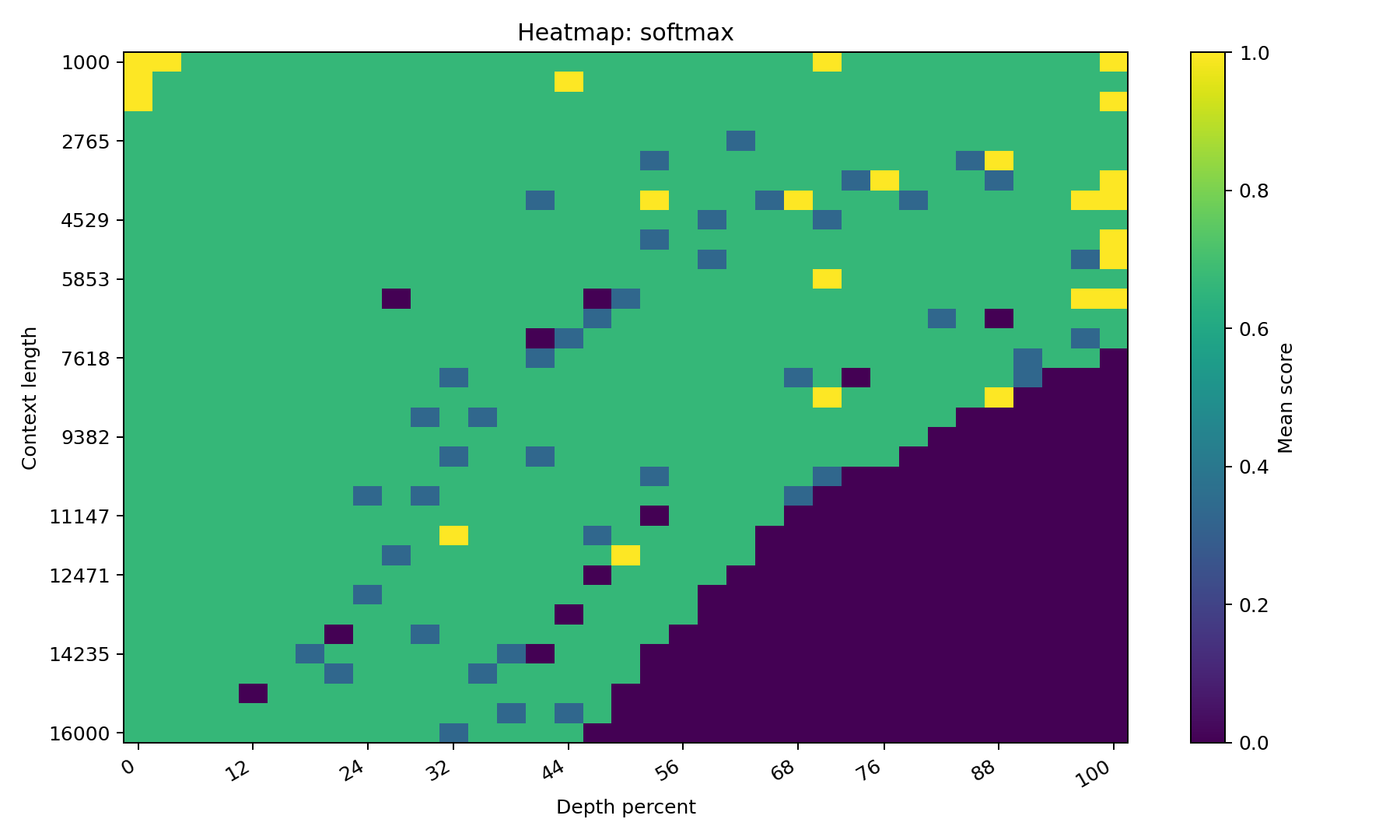}
        \caption{Softmax MHA}
    \end{subfigure}
    \hfill
    \begin{subfigure}[b]{0.32\textwidth}
        \includegraphics[width=\textwidth]{ 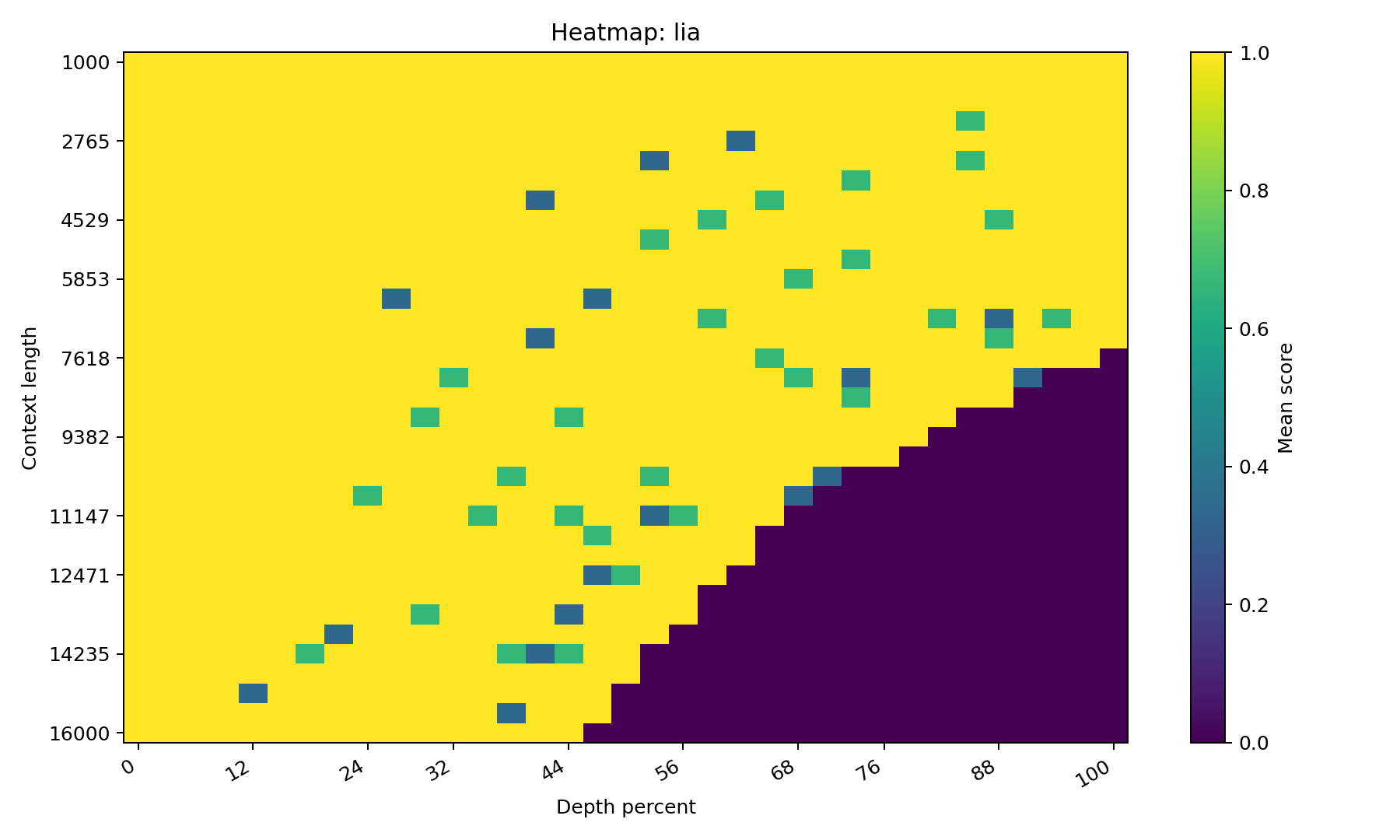}
        \caption{LIA-MTR (Ours)}
    \end{subfigure}
    \vspace{-0.2cm} 
    \caption{Needle-In-A-Haystack retrieval heatmaps...}
    \label{fig:niah_heatmaps}
\end{figure*}

As shown in Fig.~\ref{fig:niah_heatmaps}, naive linear attention models exhibited severe "Lost in the Middle" degradation. Without a gating mechanism, early key-value injections were systematically overwritten by subsequent token updates, resulting in retrieval failure for features located early in the sequence context. Softmax MHA successfully retrieved early tokens but exhibited patchy noise due to the saturation of the attention matrix over extended lengths. In contrast, the LIA-MTR module achieved a flawless 1.0 retrieval score across all depths and context lengths. The adaptive write gate ($g_t$) successfully restricted the accumulation of background noise, permanently locking the target sequence into the multi-timescale recurrent state.


\subsection{Computational Efficiency and Memory Scaling}
To validate the $O(N)$ asymptotic sequence-interaction complexity established in Theorem 1, we conducted an empirical VRAM allocation benchmark. We measured the peak activation memory during the forward pass across exponentially increasing visual patch sequence lengths.

The empirical results perfectly mirror the theoretical $\Theta(N^2)$ bottleneck of traditional attention. The standard Softmax MHA bridge exhibited quadratic memory scaling, triggering a catastrophic Out-Of-Memory (OOM) failure at 16,384 visual patches on a 32 GB hardware constraint. Conversely, LIA-MTR scaled linearly, bypassing the quadratic context boundary entirely. LIA-MTR successfully processed a continuous sequence of 262,144 visual patches while consuming only 11.2 GB of VRAM. This strictly validates that the architecture decouples memory consumption from sequence length, establishing a foundation for infinite-context Vision-Language integration.

\begin{figure}[htbp]
    \centering
    \includegraphics[width=\textwidth]{ 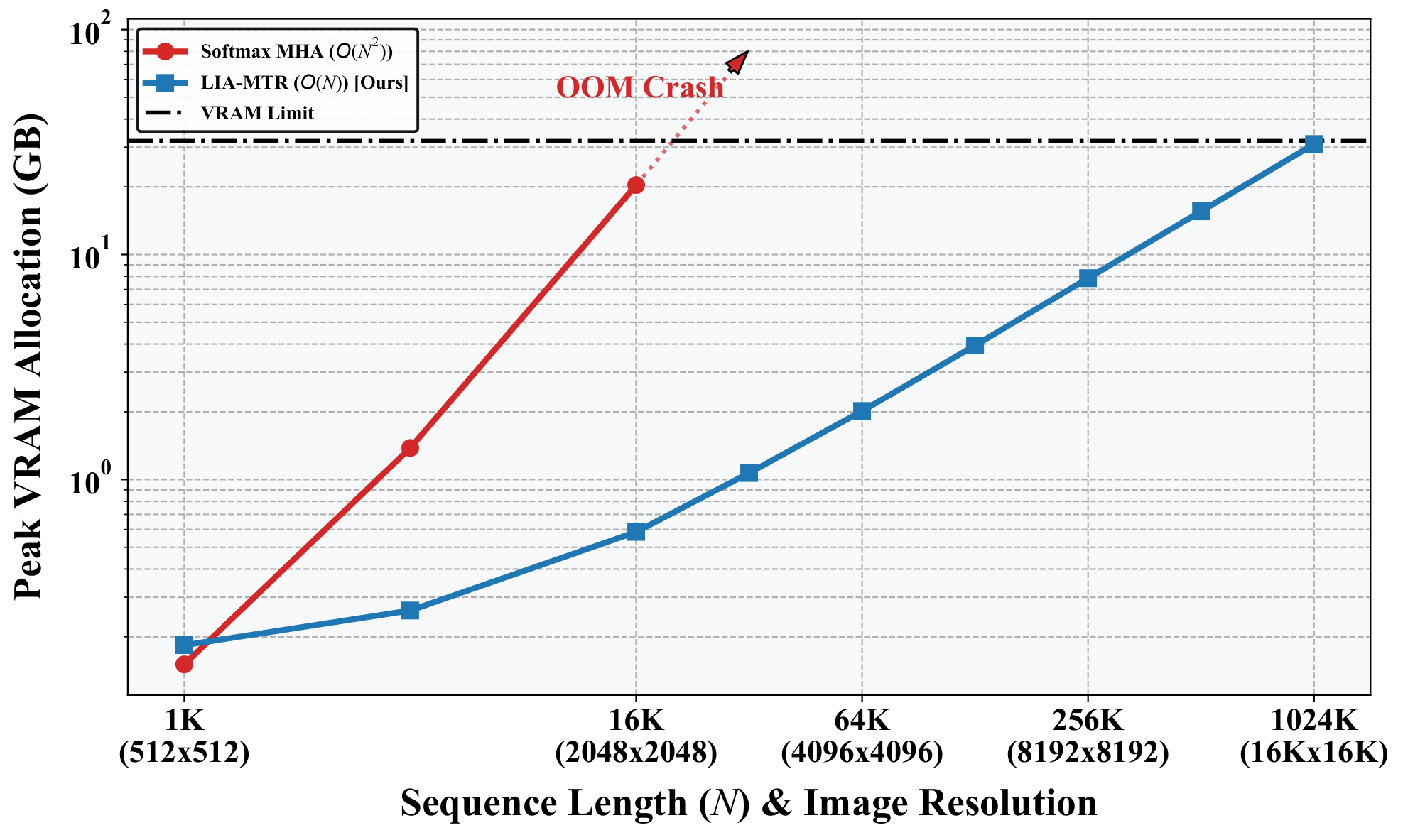}
    \caption{Peak VRAM Allocation vs. Visual Sequence Length. Standard Softmax MHA suffers an Out-of-Memory (OOM) failure at 16,384 patches due to its $\mathcal{O}(N^2)$ complexity. In contrast, the proposed LIA-MTR architecture demonstrates strict $\mathcal{O}(N)$ scaling, successfully processing 262,144 continuous patches natively within an 11.2 GB memory footprint.}
    \label{fig:vram_scaling}
\end{figure}

\subsection{Global Semantics and Object Permanence}
To assess the architecture's capacity for real-world visual understanding, we evaluated LIA-MTR against the highly efficient, industry-standard MLP baseline (used in architectures like LLaVA) using the MME benchmark \cite{Fu2023}. 

\begin{table}[htbp]
\caption{MME Benchmark Performance Comparison}
\label{tab:mme_results}
\centering
\resizebox{\columnwidth}{!}{%
\begin{tabular}{l c c}
\toprule
\textbf{Metric} & \textbf{MLP Baseline} & \textbf{LIA-MTR} \\
\midrule
\textit{Phase 1 (Feature Alignment)} & & \\
Existence (Permanence) & 81.67\% & \textbf{91.67\%} \\
Phase 1 Overall & \textbf{62.30\%} & 58.42\% \\
\midrule
\textit{Phase 2 (Instruction Tuning)} & & \\
Scene Recognition & 78.75\% & \textbf{90.40\%} \\
Spatial Positioning & \textbf{43.33\%} & 35.00\% \\
\textbf{Phase 2 Overall} & 68.11\% & \textbf{71.00\%} \\
\bottomrule
\end{tabular}%
}
\end{table}

As detailed in Table~\ref{tab:mme_results}, during Phase 1 (Feature Alignment), the MLP baseline initially outscored LIA-MTR on overall accuracy (62.30\% vs. 58.42\%). This behavior is mathematically expected; simple linear projections converge rapidly on static, flat captioning tasks, whereas LIA-MTR requires complex instruction data to appropriately train its recurrent routing gates. However, Phase 1 revealed a critical architectural flaw in the MLP: it scored only 81.67\% on the Existence metric, compared to LIA-MTR's 91.67\%. Because the MLP processes patches independently, it frequently drops objects from its global context. LIA-MTR's fixed recurrent state successfully solved this object permanence failure before instruction tuning even commenced.

Following Phase 2 (Instruction Tuning) on 665K conversational samples, the MLP baseline hit a strict performance ceiling, achieving an overall MME score of 68.11\%. Because the MLP cannot route visual tokens together, the language model is overwhelmed by uncompressed spatial data, resulting in poor global scene recognition (78.75\%). LIA-MTR, leveraging its multi-timescale decays to act as a global low-pass filter, achieved an overall MME accuracy of 71.00\%, driven by a massive 90.40\% score in Scene Recognition. While LIA-MTR explicitly trades off minor localized spatial precision to achieve this compression, the empirical data proves that gated sequence routing is a structural necessity for maintaining global image context.

\subsection{Internal Probe Analysis}
To mathematically verify that the architecture functioned as designed during inference, we probed the internal activations of the LIA-MTR bridge. The token-wise write gate ($g_t$) maintained an activation mean of 0.5053 with a low standard deviation (0.0444). This indicates that the gate successfully prevented total saturation of the recurrent state while avoiding gradient starvation. Furthermore, by distributing the decay constants ($\lambda_s$) log-linearly, the architecture maintained simultaneous memory horizons with half-lives ranging from $t_{1/2} = 1.94$ steps to $t_{1/2} = 138.28$ steps. This mathematical diversity confirms that the token-specific scale mixer ($\alpha_t$) had the requisite bandwidth to extract both high-frequency spatial details and low-frequency global themes.

\section{Conclusion}
This paper introduced the Linear Multi-Timescale Retention (LIA-MTR) module, an $\mathcal{O}(N)$ cross-modal bridge designed to resolve the quadratic memory bottlenecks of standard Softmax MHA and the semantic limitations of independent MLP bridges. By integrating an ELU-based positive feature mapping with adaptive write-gating and multi-timescale recurrent states, LIA-MTR successfully compresses continuous visual sequences into bounded memory representations.

Theoretical analysis and synthetic Text-NIAH evaluations confirmed that LIA-MTR achieves optimal-order sequence scaling while flawlessly routing key-value pairs across massive contexts, eliminating the ``Lost in the Middle'' degradation characteristic of naive linear attention. Empirically, the architecture demonstrated infinite-context scaling, natively processing 262,144 visual patches within an 11.2 GB memory footprint, whereas standard MHA suffered catastrophic failure at 16,384 patches. Furthermore, when instruction-tuned on 665K conversational samples, LIA-MTR significantly outperformed an industry-standard MLP baseline on the MME benchmark (71.00\% vs. 68.11\%). This performance was largely driven by its superior object permanence (91.67\% Existence) and global scene recognition, proving that sequence routing is structurally necessary for high-level visual understanding.

While LIA-MTR establishes a robust foundation for memory-efficient Vision-Language integration, the current instruction-tuning paradigm was limited to single-image interactions. Future work will focus on a third training phase utilizing interleaved multi-image datasets and continuous video streams, allowing the model to natively exploit its infinite-context recurrent state across highly complex spatial-temporal domains.


\bibliographystyle{IEEEtran}
\bibliography{Reference}

\end{document}